\documentclass[letterpaper]{article} 
\usepackage[preprint]{aaai2027}
\usepackage[hyphens]{url}  
\usepackage{graphicx} 
\usepackage{natbib}  
\usepackage{caption} 
\usepackage{booktabs}
\usepackage{amsmath}
\usepackage{amssymb}
\usepackage{microtype}
\usepackage{placeins}
\begin{document}

\title{MedTokenBudget: Lesion-Preserving Token Routing \\ for Dermoscopic Image Classification}

\author{Zhexiang Li}
\affiliations{
University of California, Los Angeles\\
bruceli@cs.ucla.edu
}

\maketitle

\begin{abstract}
Dermoscopy classifiers built on Vision Transformers process all image patches uniformly, although diagnostic evidence is concentrated in the lesion region. Existing token pruning methods reduce tokens using generic saliency or similarity signals, but rarely ask whether the retained subset still contains the lesion. This paper introduces MedTokenBudget, a supervised post-backbone token routing framework that learns to construct compact lesion-enriched representations when auxiliary lesion masks are available. Its Lesion-Aware Token Scoring (LATS) module fuses attention entropy, feature norm, and local feature contrast through a learned scorer, then routes the top-$K$ patches under a target budget. LATS is trained with budget curriculum learning, diversity regularization, attention distillation, and lesion-mask supervision. The trained router is evaluated with a lesion retention rate that directly measures how much ground-truth lesion evidence survives the token budget. On ISIC 2019, mask-supervised LATS consistently outperforms Random and ToMe at headline budgets while retaining substantially more lesion patches. Code is provided for reproducibility, and complete tabulated results are included in the supplementary material.
\end{abstract}

\section{Introduction}

Vision Transformers (ViTs)~\cite{dosovitskiy2021vit} and their self-supervised variants~\cite{oquab2024dinov2} have become widely used backbones for dermoscopic image classification. A ViT partitions an input image into a grid of non-overlapping patches, each projected to a token embedding, and applies self-attention over all token pairs. This design treats every spatial location as a token to be processed, even though dermoscopic evidence is spatially structured: the lesion region carries the diagnostic signal, while surrounding skin and acquisition background can occupy many patches.

A growing body of work addresses token redundancy through pruning~\cite{rao2021dynamicvit,liang2022evit,yin2022avit,kong2022spvit}, merging~\cite{bolya2023tome,ryoo2021tokenlearner}, or hybrid strategies~\cite{lee2025frequency,diffrate2023}. These methods are typically optimized for accuracy or throughput, and this is reasonable for natural-image benchmarks where the object often occupies a large fraction of the image. Dermoscopy changes the evaluation problem. A selector can preserve enough contextual signal for classification while still discarding the lesion patches that make the representation clinically meaningful. Thus, token reduction for dermoscopy should be judged not only by classification accuracy but also by whether the retained token set contains the lesion.

This evidence-preservation view exposes a gap in existing token-reduction protocols. Tokens are commonly retained based on attention scores, feature magnitudes, pairwise similarities, or learned keep/drop decisions, but the retained subset is rarely evaluated against ground-truth pathological regions. A small lesion may occupy only a few patches in a $16\times16$ grid; aggressive pruning driven by generic saliency can silently discard them. In our ISIC retention analysis, lesion-unaware selectors such as Random and NormBased preserve far less lesion area than supervised LATS under the same budget, even when their classification accuracy remains competitive. Recent work on small-lesion detection~\cite{wu2026mftmamba} and long-tail dermatology~\cite{tao2026enfusenet} further reinforces that small pathological regions remain challenging, while WSI token pruning~\cite{bozkurt2025slim,chen2026sparselearn,chen2026tcssa} addresses a different scale: selecting tiles from gigapixel slides rather than preserving lesion patches inside a standard-resolution image.

This paper proposes \textbf{MedTokenBudget}, a framework for \emph{lesion-aware} post-backbone token routing. The key premise is simple: if a token budget is used in a diagnostic representation, the selected tokens should preserve lesion evidence, not merely classifier accuracy. MedTokenBudget therefore treats lesion evidence as both a training target when masks are available and an evaluation target when retained patches can be compared with masks. The paper makes five contributions:

\begin{enumerate}
\item \textbf{A new failure mode.} We study lesion-unaware token pruning as a dermoscopic token-routing failure case: a selector can preserve classification accuracy while discarding small lesion patches.
\item \textbf{A lesion-aware routing framework.} We introduce LATS, a supervised post-backbone scorer that fuses attention entropy, feature norm, and local feature contrast to estimate patch importance before top-$K$ routing.
\item \textbf{A training strategy for compact representations.} We train the router with budget curriculum learning, diversity regularization, attention distillation, and auxiliary lesion localization, encouraging non-collapsed score maps that remain useful under aggressive budgets.
\item \textbf{A lesion retention metric.} We define lesion retention rate, the fraction of ground-truth lesion patches retained by a token selector, as a direct measure of whether token routing preserves diagnostic evidence.
\item \textbf{A systematic budget study.} We evaluate seven token selection strategies over ten budgets on ISIC 2019, include an auxiliary Kvasir v2 accuracy-only sweep, and report single-signal selector comparisons plus aligned three-seed robustness for LATS, Random, and ToMe at the headline budgets.
\end{enumerate}

\section{Related Work}

\subsection{Token Pruning and Merging}

The quadratic complexity of self-attention~\cite{vaswani2017attention} has motivated token reduction through pruning~\cite{rao2021dynamicvit,liang2022evit,kong2022spvit,yin2022avit}, merging~\cite{bolya2023tome,ryoo2021tokenlearner}, and hybrid compression~\cite{diffrate2023,lee2025frequency}. These methods rank or merge tokens using class-token attention, learned keep/drop masks, feature similarity, or frequency cues. Their evaluation usually centers on accuracy-throughput tradeoffs. MedTokenBudget studies a different axis: whether a retained token subset preserves the pathological region that makes the prediction interpretable as lesion-based evidence.

\subsection{Token Efficiency and MIL in Medical Imaging}

Medical token efficiency has focused mostly on MIL for gigapixel WSIs~\cite{ilse2018attention,lu2021clam,shao2021transmil} and recent WSI pruning systems~\cite{bozkurt2025slim,chen2026sparselearn,chen2026tcssa}. These operate at a much coarser spatial scale---selecting tiles from slides---than patch selection within a single dermoscopic image. MedSpaformer~\cite{ye2026medspaformer}, HeartLLM~\cite{yang2026heartllm}, and CheXficient~\cite{wang2026chexficient} address adjacent efficient medical modeling, while PrATo~\cite{dutta2025prato} studies prompt-guided token pruning. Prompt-guided and attribution-based methods answer a related but different localization question: they use external spatial prompts or prediction-conditioned relevance maps to explain or guide an already formed representation. MedTokenBudget instead learns a prompt-free forward-pass router for classification, using masks only as auxiliary supervision when available and evaluating the retained token set directly against lesion masks.

\section{Method}

\subsection{Overview}

\begin{figure*}[t]
\centering
\includegraphics[width=0.95\textwidth]{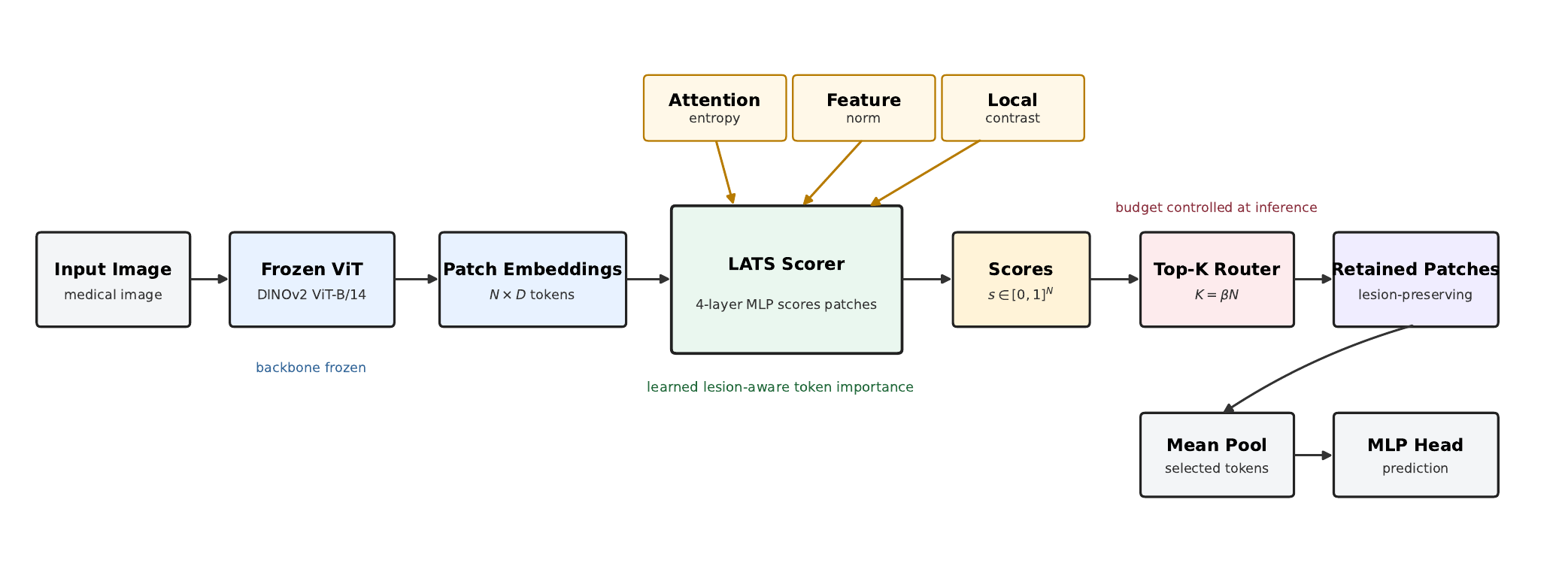}
\caption{MedTokenBudget overview. A frozen ViT produces patch embeddings; LATS scores patches with attention, norm, and local-contrast signals; the top-$K$ router keeps a budgeted subset for MLP classification.}
\label{fig:overview}
\end{figure*}

MedTokenBudget is a post-backbone routing framework that converts a full ViT patch sequence into a compact lesion-enriched token subset (Figure~\ref{fig:overview}). The framework deliberately leaves the frozen backbone unchanged: it does not claim backbone acceleration, but instead studies which patch embeddings remain available to downstream classifiers or token-consuming modules after routing. This decoupled design makes the selection policy explicit and allows lesion retention to be evaluated directly.

Given an input image $\mathbf{x} \in \mathbb{R}^{H \times W \times 3}$, the frozen backbone $\mathcal{B}$ extracts patch embeddings $\mathbf{Z} = \mathcal{B}(\mathbf{x}) \in \mathbb{R}^{N \times D}$, where $N = 256$ for a ViT-B/14 at resolution $224\times224$ and $D = 768$. Let $[N]=\{1,\ldots,N\}$. The LATS module produces per-patch importance scores $\mathbf{s} \in [0,1]^N$. A top-$K$ router selects the $K = \max(K_{\min}, \lfloor \beta N \rfloor)$ highest-scoring patches with $K_{\min}=16$, where $\beta \in (0,1]$ is the token budget. The retained patches are mean-pooled and passed to an MLP head $\mathcal{H}$ for classification. The following subsections describe the scorer signals, differentiable routing and curriculum, and auxiliary training losses.

\subsection{Lesion-Aware Token Scoring (LATS)}

LATS computes a lesion-relevance score for each patch by combining patch embeddings and auxiliary signals through a learned MLP $\mathcal{S}_\theta$. The design uses three weak but complementary cues: attention concentration, semantic foreground strength, and local feature transitions. None is assumed to be sufficient alone; the scorer learns how to combine them under mask-supervised and classification-supervised training.

\noindent\textbf{Attention Entropy.} Patches receiving \emph{focused} attention from the ViT's self-attention are more likely to contain discriminative features. Let $\alpha_{ij} = \frac{1}{n_h}\sum_{h=1}^{n_h} \alpha_{ij}^{(h)}$ be the mean attention weight from patch $i$ to patch $j$ in the last transformer block, averaged over $n_h$ heads. The attention entropy for patch $i$ is:
\begin{equation}
H_i = -\sum_{j=1}^{N} \alpha_{ij} \log \alpha_{ij}, \quad
s_i^{\text{attn}} = 1 - \frac{H_i}{\log N}
\end{equation}
Low entropy (concentrated attention) yields a high score. In the reported DINOv2 experiments, last-block attention is reconstructed from the frozen backbone's query-key projections and averaged over heads. Cosine similarity between patch features is implemented only as a pseudo-attention fallback for backbones that do not expose query-key attention.

\noindent\textbf{Feature Norm.} DINOv2 patch token norms correlate with semantic content: background patches exhibit lower norms than foreground regions. Per-image normalized scores are computed as:
\begin{equation}
s_i^{\text{norm}} = \frac{\|\mathbf{z}_i\|_2}{\max_{j \in [N]} \|\mathbf{z}_j\|_2}
\end{equation}

\noindent\textbf{Local Feature Contrast.} Lesion boundaries produce sharp transitions in feature space where neighboring patches exhibit high feature dissimilarity. Local contrast is estimated via cosine dissimilarity with adjacent patches in the rasterized 1D token sequence. Each adjacent-pair dissimilarity is assigned to both participating patches and averaged over available neighbors, then used to gate a learned linear projection $\mathbf{W}_{\text{freq}}$:
\begin{align}
\delta_i &= 1 - \frac{\mathbf{z}_i^\top \mathbf{z}_{i+1}}{\|\mathbf{z}_i\|_2 \|\mathbf{z}_{i+1}\|_2}, \\
c_i &= \frac{\mathbb{1}_{i>1}\delta_{i-1} + \mathbb{1}_{i<N}\delta_i}{\mathbb{1}_{i>1}+\mathbb{1}_{i<N}}, \\
\mathbf{f}_i^{\text{contrast}} &= c_i\,\mathbf{W}_{\text{freq}}\mathbf{z}_i .
\end{align}
The learned projection $\mathbf{W}_{\text{freq}}$ maps each patch embedding to a reduced space (dimension $D/16$) where it is gated by the scalar local dissimilarity score, up-weighting features at boundary-like transitions. The 1D raster adjacency is an implementation simplification: it ignores vertical neighbors and treats row-boundary pairs as adjacent, so the local-contrast signal should be read as a lightweight raster feature rather than a full 2D boundary detector. A 2D neighborhood contrast variant remains future work.

\noindent\textbf{Scoring MLP.} The auxiliary signals are concatenated with $\mathbf{z}_i$ and processed by a 4-layer MLP (256 hidden, LayerNorm, GELU, dropout 0.1) with a sigmoid output: $s_i = \mathcal{S}_\theta([\mathbf{z}_i; s_i^{\text{attn}}; s_i^{\text{norm}}; \mathbf{f}_i^{\text{contrast}}])$.

\subsection{Token Router and Budget Curriculum}

The router converts continuous scores into a fixed token budget. During training, differentiable top-$K$ selection uses Gumbel-perturbed scores~\cite{jang2017gumbel} with temperature $\tau = 0.5$. Let $\ell_i=\operatorname{logit}(s_i)$ and $\tilde{\ell}_i=(\ell_i+g_i)/\tau$, where $g_i$ is Gumbel noise. The forward pass uses a hard top-$K$ mask $\mathbf{m}^{\mathrm{hard}}$, while the backward pass uses the straight-through relaxation $\mathbf{m}=\mathbf{m}^{\mathrm{hard}}+\mathbf{m}^{\mathrm{soft}}-\operatorname{sg}(\mathbf{m}^{\mathrm{soft}})$ with $\mathbf{m}^{\mathrm{soft}}=\operatorname{clip}(K\operatorname{softmax}(\tilde{\ell}),0,1)$. During training the straight-through mask is applied to the full patch matrix and the masked sum is divided by the mask mass; at inference, deterministic hard top-$K$ is applied before pooling. Spatial smoothing (3$\times$3 average pooling on the 2D score grid) encourages contiguous retention of lesion regions.

The token budget $\beta$ follows a cosine-annealed schedule: $\beta$ decreases from $0.5$ to $0.25$ over 20 epochs, after which it remains at $0.25$ for the remainder of training. This curriculum forces the scorer to learn meaningful rankings under moderate budgets early, before facing aggressive pruning.

\subsection{Training Objectives}

The training objective combines classification with auxiliary scorer losses. The implementation also records a budget-deviation term for accounting, but the trainable scorer signal comes from classification, diversity, lesion localization, and attention distillation.

\noindent\textbf{Classification.} Standard cross-entropy $\mathcal{L}_{\text{cls}}$ on the MLP head's predictions from the retained patches.

\noindent\textbf{Budget Deviation Diagnostic.} The reported implementation fixes $K=\max(K_{\min},\lfloor\beta N\rfloor)$ before top-$K$ selection, so $\mathcal{L}_{\text{budget}}=|K/N-\beta|$ is a deterministic diagnostic rather than a scorer-training signal. It records deviation introduced by integer rounding and the minimum-token constraint. The scalar loss includes this term with weight $0.01$, but its gradient with respect to the scorer and classifier is zero once $\beta$ is fixed.

\noindent\textbf{Diversity Regularization.} To prevent the scorer from collapsing to a fixed spatial pattern across all images, pairwise cosine similarity between score vectors of different images within a batch is penalized:
\begin{equation}
\mathcal{L}_{\text{div}} = \frac{1}{B(B-1)} \sum_{i \neq j} \max(0, \cos(\mathbf{s}^{(i)}, \mathbf{s}^{(j)}))
\end{equation}

\noindent\textbf{Lesion Localization.} When ground-truth segmentation masks are available, a binary cross-entropy loss encourages the scorer to predict high scores on lesion patches. Let $m_i \in \{0,1\}$ denote the patch-level lesion mask:
\begin{equation}
\mathcal{L}_{\text{lesion}} = -\frac{1}{N}\sum_{i=1}^{N}
\left[m_i \log s_i + (1-m_i)\log(1-s_i)\right]
\end{equation}
On ISIC, lesion masks from ISIC 2018 serve as auxiliary training supervision when a training image has a valid overlapping mask.

\noindent\textbf{Attention Distillation.} The backbone's self-attention weights provide a teacher signal for patch importance. The CLS-to-patch attention map is max-normalized and matched to the scorer outputs via MSE:
\begin{equation}
\mathcal{L}_{\text{distill}} = \frac{1}{N} \sum_{i=1}^{N} \left(s_i - \frac{\bar{\alpha}_{\text{CLS}\rightarrow i}}{\max_j \bar{\alpha}_{\text{CLS}\rightarrow j}}\right)^2
\end{equation}
where $\bar{\alpha}_{\text{CLS}\rightarrow i} = \frac{1}{n_h}\sum_{h=1}^{n_h} \alpha^{(h)}_{\text{CLS}\rightarrow i}$ is the mean attention from the class token to patch $i$.

The implemented scalar objective is $\mathcal{L}_{\text{train}} = \mathcal{L}_{\text{cls}} + 0.01\mathcal{L}_{\text{budget}} + 0.05\mathcal{L}_{\text{div}} + 0.1\mathcal{L}_{\text{lesion}} + 0.1\mathcal{L}_{\text{distill}}$. Since $\mathcal{L}_{\text{budget}}$ is deterministic under fixed $K$, it is reported as a diagnostic of rounding and minimum-token constraints, not as a meaningful scorer-training signal.

\section{Experimental Setup}

\subsection{Datasets}

The primary benchmark is \textbf{ISIC 2019}~\cite{tschandl2020human}, which contains 25,331 dermoscopic images across 8 skin lesion categories. We use an 80/20 random split with seed 42, giving 5,067 validation images in the reported split. Lesion sizes vary substantially across images, making ISIC a suitable testbed for whether token routing preserves localized diagnostic evidence. ISIC 2018 segmentation masks serve as auxiliary supervision for overlapping training images and as evaluation masks for retention. In the actual experiment split, 617 training images and 155 validation images overlap with valid ISIC 2018 masks; these masks cover MEL/NV cases only, so retention metrics are a focused lesion-overlap analysis rather than a full-class validation metric. After resizing masks to the $16\times16$ patch grid, lesion-positive patches occupy 31.6\% of the mask-overlap validation patches on average. Validation masks are never used for training, early stopping, or hyperparameter selection. Because the reported split is image-level rather than patient/lesion-grouped, we interpret classification accuracy as benchmark evidence rather than a patient-level generalization guarantee.

An auxiliary accuracy-only sweep on Kvasir v2~\cite{pogorelov2017kvasir} (8,000 endoscopic images, 8 classes, no masks) is reported in the supplementary material as an implementation check outside dermoscopy.

\subsection{Implementation}

DINOv2 ViT-B/14~\cite{oquab2024dinov2} serves as the frozen backbone ($N = 256$ patches, $D = 768$) for the reported experiments. The implementation extracts last-block patch attention directly from DINOv2's frozen query-key projections for both attention entropy and CLS-attention distillation. The LATS scorer is a 4-layer MLP (256 hidden; GELU; LayerNorm); the classifier head is a 2-layer MLP (256$\rightarrow$128 hidden). Total trainable parameters are $\sim$0.74M versus 86M frozen. Training runs 50 epochs with AdamW (lr $= 10^{-3}$, weight decay $= 10^{-5}$), cosine LR, batch size 64, and early stopping patience 12. One NVIDIA A100 (80GB) completes ISIC training in approximately 2 hours.

\subsection{Baselines and Evaluation}

The evaluation asks whether LATS improves low-budget classification, preserves lesion evidence better than lesion-unaware selection, and which design choices explain low-budget behavior. Seven token selection strategies are compared. Each non-LATS baseline trains its own independent MLP head (identical architecture) for 30 epochs on the token distribution produced by that method. For budget-sweep baselines, the head is trained with budgets sampled from the sweep range; NoPruning uses the full-token budget. This makes baseline heads robust across budgets but is not fully epoch-matched to LATS, which receives 50 epochs because it jointly learns the scorer and head. The methods are NoPruning, Random, NormBased, ToMe~\cite{bolya2023tome}, AttentionEntropy, LocalContrast, and supervised LATS.

All methods are evaluated at ten budgets $\beta \in \{0.1, 0.2, \ldots, 1.0\}$. Metrics include classification accuracy, macro F1, balanced accuracy, and where masks are available, \emph{lesion retention rate}: the fraction of ground-truth lesion patches retained in the top-$K$. ISIC masks are resized to the $16\times16$ patch grid, thresholded at $>0.5$ to form binary patch labels, flattened to token order, and evaluated only for validation images with valid overlapping masks. LATS uses auxiliary lesion-mask supervision when overlapping masks exist, while Random and heuristic selectors do not; retention comparisons therefore evaluate the supervised LATS method rather than an unsupervised selector-only setting. Main sweep tables report a single training run and one Random draw unless otherwise specified; a three-seed ISIC robustness check (seeds 42, 43, 44) for LATS, Random, and ToMe at $\beta \in \{0.1,0.3,0.5\}$ is reported in the main results. The single-signal analysis uses independently trained selector baselines from the budget sweep as training-time controls.

\section{Results}

\subsection{Main Results on ISIC 2019}

Table~\ref{tab:main} reports accuracy and lesion retention for all methods at representative budgets on ISIC 2019. Full ten-budget results appear in the supplementary material.

\begin{table*}[t]
\centering
\begin{tabular*}{\textwidth}{@{\extracolsep{\fill}}lccccc|c}
\toprule
Method & $\beta{=}0.1$ & $\beta{=}0.2$ & $\beta{=}0.3$ & $\beta{=}0.5$ & $\beta{=}1.0$ & Ret@0.3 \\
\midrule
NoPruning & 77.5 & 77.5 & 77.5 & 77.5 & 77.5 & 100.0 \\
Random    & 76.2 & 77.0 & 76.8 & 77.1 & \textbf{77.0} & 29.5 \\
NormBased & 69.7 & 71.8 & 73.1 & 74.5 & 75.7 & 21.1 \\
ToMe      & 75.2 & 76.5 & 76.9 & 77.0 & \textbf{77.0} & --- \\
AttnEnt   & 70.9 & 72.9 & 73.9 & 75.2 & 76.8 & 34.0 \\
LocalCtr  & 73.9 & 74.8 & 75.7 & 75.8 & 76.5 & 19.3 \\
\textbf{LATS} & \textbf{77.5} & \textbf{80.1} & \textbf{80.2} & \textbf{79.3} & 75.1 & \textbf{67.4} \\
\bottomrule
\end{tabular*}
\caption{Single-run classification accuracy (\%) and lesion retention (\%) at selected budgets on ISIC 2019. \textbf{Bold}: best per budget excluding NoPruning, which uses all tokens. Ret@0.3 is a representative retention column at $\beta{=}0.3$ on the 155-image mask-overlap subset; full retention curves appear in Figure~\ref{fig:retention}. ToMe retention is unavailable because merged tokens do not preserve one-to-one patch masks. Random uses one draw. Three-seed results appear in Table~\ref{tab:robust}.}
\label{tab:main}
\end{table*}

\begin{table*}[t]
\centering
\begin{tabular*}{\textwidth}{@{\extracolsep{\fill}}llcccc}
\toprule
$\beta$ & Method & Acc. (\%) & Macro F1 (\%) & Bal. Acc. (\%) & Ret. (\%) \\
\midrule
0.1 & LATS & $77.77{\pm}0.27$ & $67.08{\pm}0.25$ & $65.89{\pm}0.49$ & $28.9{\pm}0.1$ \\
0.1 & Random & $75.75{\pm}0.44$ & $59.50{\pm}1.34$ & $55.39{\pm}1.31$ & $9.9{\pm}0.1$ \\
0.1 & ToMe & $75.94{\pm}0.61$ & $60.96{\pm}1.56$ & $56.90{\pm}1.57$ & --- \\
0.3 & LATS & $80.38{\pm}0.18$ & $69.17{\pm}1.28$ & $66.40{\pm}0.91$ & $67.4{\pm}0.2$ \\
0.3 & Random & $76.57{\pm}0.30$ & $60.95{\pm}1.06$ & $56.72{\pm}1.81$ & $29.7{\pm}0.3$ \\
0.3 & ToMe & $76.93{\pm}0.03$ & $61.92{\pm}0.23$ & $57.81{\pm}0.48$ & --- \\
0.5 & LATS & $79.78{\pm}0.51$ & $67.61{\pm}1.83$ & $64.14{\pm}1.67$ & $85.8{\pm}0.2$ \\
0.5 & Random & $76.74{\pm}0.35$ & $61.68{\pm}0.72$ & $57.15{\pm}1.20$ & $49.9{\pm}0.2$ \\
0.5 & ToMe & $77.13{\pm}0.37$ & $62.82{\pm}0.77$ & $58.73{\pm}0.95$ & --- \\
\midrule
\multicolumn{6}{l}{\footnotesize LATS (mask-free, same architecture/epochs, $\mathcal{L}_{\text{lesion}}$ disabled)} \\
0.1 & LATS$_{\text{mf}}$ & $66.5{\pm}3.0$ & $51.2{\pm}3.1$ & $51.6{\pm}4.2$ & $19.5{\pm}14.2$ \\
0.3 & LATS$_{\text{mf}}$ & $73.2{\pm}1.9$ & $55.9{\pm}2.7$ & $54.3{\pm}2.3$ & $46.6{\pm}29.1$ \\
0.5 & LATS$_{\text{mf}}$ & $74.8{\pm}0.8$ & $56.7{\pm}2.7$ & $53.9{\pm}2.3$ & $62.7{\pm}29.4$ \\
\bottomrule
\end{tabular*}
\caption{Aligned three-seed ISIC robustness for the headline methods and budgets. Retention is computed on the 155-image mask-overlap validation subset for methods with one-to-one retained patches; ToMe merges tokens, so patch retention is not reported.}
\label{tab:robust}
\end{table*}

Table~\ref{tab:robust} additionally includes LATS$_{\text{mf}}$ (same LATS architecture and training budget, but with $\mathcal{L}_{\text{lesion}}$ disabled), which separates the supervised contribution from the routing architecture. Removing mask supervision reduces both accuracy and retention, as expected. The mean mask-free retention remains above Random at the three headline budgets ($19.5\%$ vs.\ $9.9\%$ at $\beta{=}0.1$, $46.6\%$ vs.\ $29.7\%$ at $\beta{=}0.3$, $62.7\%$ vs.\ $49.9\%$ at $\beta{=}0.5$), but the seed-to-seed range is wide: 3.1--28.5\%, 13.1--66.0\%, and 29.3--85.0\% retention at $\beta=0.1,0.3,0.5$, respectively. We therefore treat LATS$_{\text{mf}}$ as a diagnostic control rather than a stable replacement for mask supervision. The lower accuracy at $\beta{=}0.1$ ($66.5{\pm}3.0\%$ vs.\ Random's $75.8{\pm}0.4\%$) further indicates that the scorer requires either mask supervision or a larger token budget in the most aggressive regime.

The ISIC results support three observations. First, supervised lesion-aware routing is most useful in the low- to mid-budget regime. In the single-run sweep, LATS reaches 80.2\% accuracy at $\beta=0.3$, compared with 77.5\% for the full-token NoPruning reference. Because NoPruning is single-run, the stronger comparison is Table~\ref{tab:robust}: at $\beta=0.3$, LATS obtains $80.38{\pm}0.18\%$ accuracy, outperforming Random ($76.57{\pm}0.30\%$) and ToMe ($76.93{\pm}0.03\%$) under the same three seeds. Second, LATS also improves macro F1 and balanced accuracy over Random and ToMe at all three headline budgets. Third, LATS is not intended to be a full-token classifier; its accuracy peaks at intermediate budgets, consistent with curriculum specialization for compact representations.

\subsection{Lesion Retention}

\begin{figure}[tbp]
\centering
\includegraphics[width=\linewidth]{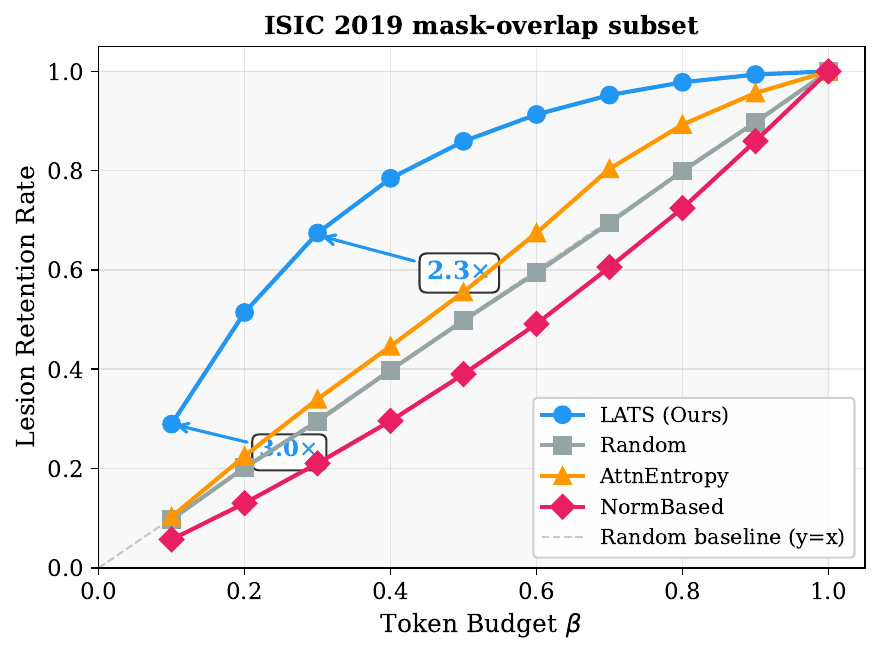}
\caption{Lesion retention vs.\ token budget on ISIC 2019.}
\label{fig:retention}
\end{figure}

The retention metric explains why the same accuracy table can hide qualitatively different token subsets. Retention is recall-like: it measures how much lesion area remains under a fixed token budget, while the fixed $K$ controls the number of retained background patches. The supervised LATS model is substantially stronger than lesion-unaware selection: at $\beta=0.3$, LATS retains 67.4\% of lesion patches in the single-run sweep, while Random retains 29.5\%. The aligned three-seed check gives the same conclusion: $67.4{\pm}0.2\%$ for LATS versus $29.7{\pm}0.3\%$ for Random, with paired seed-wise retention gains of 37.5--37.9 percentage points. At $\beta=0.1$, the relative gap is larger ($28.9{\pm}0.1\%$ vs.\ $9.9{\pm}0.1\%$), while at $\beta=0.5$ the absolute gap remains large ($85.8{\pm}0.2\%$ vs.\ $49.9{\pm}0.2\%$). Because LATS uses auxiliary mask supervision when available and Random has no trainable scorer or mask access, this is not a mask-symmetric comparison; it shows the effect of training a lesion-aware router to construct compact representations. NormBased retains only 21.1\% of lesion patches at $\beta=0.3$, showing that feature magnitude alone is not a reliable proxy for lesion evidence in these dermoscopic images. Figure~\ref{fig:retention} plots the full retention curve.

This is the central empirical point: classification accuracy alone does not reveal whether the retained representation still contains the lesion. Random selection reaches 76.8\% accuracy at $\beta=0.3$ in the main sweep, but it preserves less than half as much lesion area as supervised LATS on the mask-overlap subset. For downstream modules that consume selected tokens, the classifier may remain competitive while the representation has discarded much of the visible lesion evidence.

\begin{figure*}[t]
\centering
\includegraphics[width=\textwidth]{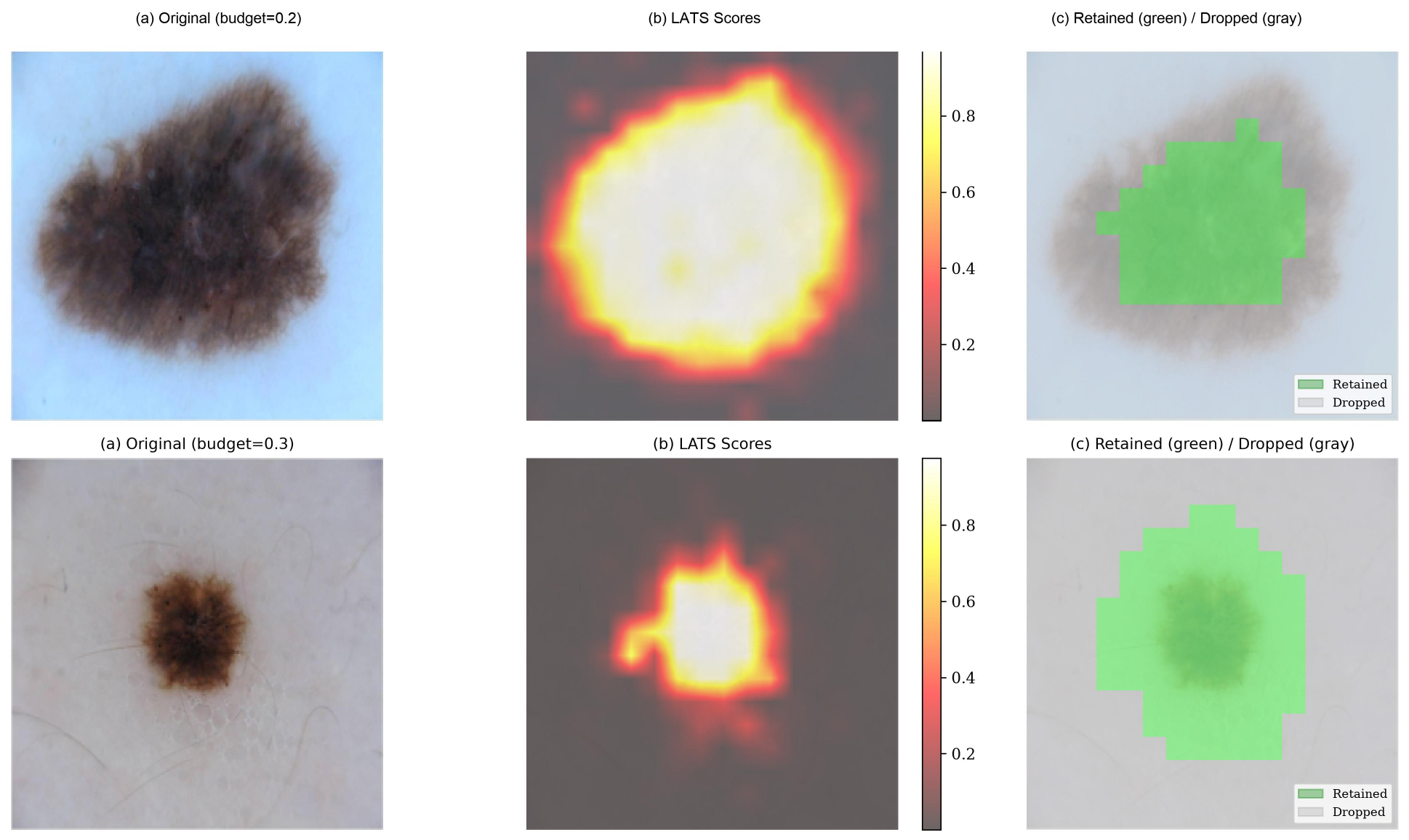}
\caption{Qualitative routing examples. LATS assigns high scores to lesion regions and retains compact patch subsets concentrated over visible lesion evidence.}
\label{fig:qualitative}
\end{figure*}

Figure~\ref{fig:qualitative} provides a qualitative example; the main evidence remains the quantitative retention in Table~\ref{tab:main} and Figure~\ref{fig:retention}.

\FloatBarrier
\section{Analysis}

\subsection{Single-Signal Selector Baseline Comparison}

Each single-signal selector baseline (AttentionEntropy, NormBased, LocalContrast) isolates one LATS scoring signal while using an identical head architecture and training budget in the budget sweep. Table~\ref{tab:signal} compares these independently trained selector baselines against the full LATS model at $\beta=0.3$ on ISIC; this is distinct from the inference-time signal-toggle ablation in the result artifacts.

\begin{table}[tbp]
\centering
\begin{tabular}{lccc}
\toprule
Method & Acc. (\%) & F1 (\%) & Ret. (\%) \\
\midrule
LATS & \textbf{80.2} & \textbf{68.0} & \textbf{67.4} \\
AttnEnt & 73.9 & 57.3 & 34.0 \\
NormBased & 73.1 & 54.3 & 21.1 \\
LocalContrast & 75.7 & 57.1 & 19.3 \\
\bottomrule
\end{tabular}
\caption{Single-run single-signal selector baseline comparison at $\beta=0.3$ on ISIC 2019. Each baseline uses one LATS signal in isolation with an independently trained head; values come from the budget-sweep baseline runs.}
\label{tab:signal}
\end{table}

The full LATS model outperforms the independently trained single-signal selector baselines by 4.5--7.1 accuracy points and 2--3.5$\times$ in retention in this single-run comparison. The contrast between accuracy and retention is informative: LocalContrast is the strongest single-signal classifier, yet it has the lowest retention. Boundary-like changes can therefore help classification without reliably preserving lesion interiors.

\subsection{Budget Curriculum Ablation}

To quantify the effect of budget curriculum training, a fixed-budget variant is trained with $\beta = 0.5$ held constant. Table~\ref{tab:curriculum} compares the two training strategies at selected budgets.

\begin{table*}[t]
\centering
\begin{tabular*}{\textwidth}{@{\extracolsep{\fill}}lccccc}
\toprule
Training & $\beta{=}0.1$ & $\beta{=}0.2$ & $\beta{=}0.3$ & $\beta{=}0.5$ & $\beta{=}1.0$ \\
\midrule
Curriculum & 77.5 & 80.1 & 80.2 & 79.3 & 75.1 \\
Fixed ($\beta{=}0.5$) & 72.4 & 77.4 & 79.0 & 80.0 & 76.4 \\
\bottomrule
\end{tabular*}
\caption{Single-run curriculum vs.\ fixed-budget training on ISIC 2019.}
\label{tab:curriculum}
\end{table*}

The curriculum model is better under aggressive budgets, improving over fixed-budget training by 5.1 points at $\beta=0.1$ and 2.7 points at $\beta=0.2$. Since this is a single-run ablation, we interpret it as diagnostic evidence rather than a definitive causal estimate. The fixed-budget model is slightly better at $\beta=1.0$, where neither model is operating in the target compressed regime.

\subsection{Random Competitiveness and Why It Does Not Suffice}

Random selection remains a strong classifier on ISIC, especially at moderate and high budgets. The relevant distinction is therefore not whether Random collapses, but whether it preserves the right evidence. The three-seed comparison shows that LATS improves over both Random and ToMe at $\beta=0.1,0.3,0.5$; more importantly, it retains $2.9\times$, $2.3\times$, and $1.7\times$ as many lesion patches as Random at the same budgets. Thus, the contribution is a token subset more aligned with the visible lesion under the same budget.

\FloatBarrier
\section{Discussion}

\subsection{Why Lesion Retention Matters}

The experiments highlight a gap between predictive performance and evidence preservation. A classifier can exploit background skin texture, acquisition patterns, or class-correlated context and still perform well, even if the retained token subset omits part of the lesion. Lesion retention measures whether the compact representation still contains the visible lesion region. We view it as a representation-level evidence-preservation proxy, not as a standalone clinical reliability guarantee.

\subsection{Efficiency Characteristics}

MedTokenBudget routes tokens after feature extraction, so it does not reduce frozen-backbone latency. Its efficiency target is representational compression: reducing 256 patch embeddings to a smaller lesion-enriched set for downstream heads or token-consuming modules. This design is deliberate. In-backbone routing can provide wall-clock speedups, but it couples feature extraction to a specific pruning policy. Post-backbone routing preserves a shared frozen feature extractor and allows different downstream tasks to use different scorers and budgets. The results should therefore be read as evidence for compact evidence-preserving representations, not as a claim of end-to-end inference acceleration.

\subsection{Limitations}

The scope is intentionally narrow. The strongest evidence is dermoscopy-specific, and non-dermoscopic generalization remains open because auxiliary datasets without lesion masks cannot test the central evidence-preservation claim. The framework is evaluated with DINOv2 as the frozen backbone; other foundation models and in-backbone routing systems should be studied separately. The ISIC retention analysis uses the 155 validation images that overlap with valid ISIC 2018 segmentation masks. This subset is small and class-limited (MEL/NV), and if such overlap images are biased toward larger or easier lesions, retention estimates could overstate evidence preservation. Larger mask-annotated evaluations, lesion-size stratification, and patient/lesion-grouped splits would strengthen the evidence.

A mask-free LATS control (same architecture and training budget, $\mathcal{L}_{\text{lesion}}$ disabled) is included in Table~\ref{tab:robust}; its higher mean retention but high seed variance suggests that the routing architecture can help, while stable low-budget behavior depends on mask supervision. Remaining controls---a mask-oracle upper bound, fully epoch-matched baseline retraining, attribution-based top-$K$ selectors such as GradCAM or transformer relevance, and formal significance testing over seeds---would further sharpen attribution and are left to future work. Finally, three-seed robustness is reported for the ISIC headline budgets, but complete ten-budget sweeps, curriculum ablation, stratified re-splits, and formal significance testing remain future work.

\section{Conclusion}

This paper introduced MedTokenBudget, a supervised framework for lesion-preserving post-backbone token routing in dermoscopic image classification. The central claim is deliberately focused: under a fixed token budget, compact representations should retain lesion evidence, not only maintain classifier accuracy. LATS operationalizes this goal with a learned multi-signal scorer, budget curriculum training, regularization, distillation, and auxiliary mask supervision when masks are available. On ISIC, LATS consistently improves over Random and ToMe at the headline budgets and retains $1.7$--$2.9\times$ more lesion patches than Random across $\beta=0.1,0.3,0.5$. These results position lesion retention as an important evaluation dimension for medical token routing and provide a controlled dermoscopic benchmark for future token-selection methods.

\section*{Ethical Statement}

All datasets are publicly available: ISIC 2019 (CC-0/CC-BY), ISIC 2018 (CC-0), and Kvasir v2 (restricted to research and educational use; other uses, including commercial use, require prior written permission from the dataset owners). The study uses de-identified public data only and collects no new human-subject data. For reproducibility, the anonymous code package includes training/evaluation scripts, configuration files, and command-line instructions; complete numeric results are tabulated in the paper and supplementary material. AI tools were used for code and manuscript editing assistance; all reported experiments, tables, and claims were checked by the authors.

\FloatBarrier
\bibliography{references}

\clearpage
\appendix
\section*{Supplementary Material}

\subsection{Training Configuration}

Table~\ref{tab:supp_training_config} summarizes the shared backbone, optimizer, curriculum, and loss settings used for the reported runs. Baseline selectors use the same classifier-head architecture unless otherwise noted.

\begin{center}
\centering
\small
\begin{tabular}{@{}p{0.35\linewidth}p{0.58\linewidth}@{}}
\toprule
Parameter & Value \\
\midrule
Backbone & DINOv2 ViT-B/14 (frozen, 86M params) \\
Patch grid & $16\times16$ ($N = 256$) \\
Embedding dimension $D$ & 768 \\
LATS scorer & 4-layer MLP, 256 hidden, GELU, LayerNorm \\
Classifier head & 2-layer MLP, 256$\rightarrow$128 hidden \\
Total trainable & $\sim$0.74M \\
Optimizer & AdamW (lr $= 10^{-3}$, wd $= 10^{-5}$) \\
LR schedule & Cosine to 0 over 50 epochs \\
Batch size & 64 \\
Epochs & 50 (early stop patience 12) \\
Budget curriculum & Cosine, $\beta: 0.5 \to 0.25$ over 20 epochs \\
Gumbel temperature $\tau$ & 0.5 \\
Loss weights & $\lambda_{\text{div}}{=}0.05$, $\lambda_{\text{lesion}}{=}0.1$, $\lambda_{\text{distill}}{=}0.1$ \\
Budget diagnostic & weight $0.01$ on fixed-$K$ rounding/min-token deviation; no scorer gradient \\
Baseline head epochs & 30 \\
Training time (ISIC) & $\sim$2 h on 1$\times$A100 80GB \\
\bottomrule
\end{tabular}
\captionof{table}{Full training hyperparameters.}
\label{tab:supp_training_config}
\end{center}

Local contrast is implemented on the rasterized 1D token order rather than a 2D grid neighborhood. This choice keeps the selector independent of backbone-specific patch layouts and matches the forward-pass baseline implementation; replacing it with an explicit 2D neighborhood is a natural extension for future work.

\subsection{Algorithmic and Reproducibility Details}

The anonymous code package includes the training/evaluation source code, configuration files, dataset loaders, and command-line recipes needed to reproduce the reported runs. Full ISIC runs used a single NVIDIA A100 80GB GPU with PyTorch 2.1 or newer, torchvision 0.16 or newer, and \texttt{timm==1.0.15}. The main ISIC robustness table uses seeds 42, 43, and 44. Complete numeric results are tabulated in the main paper and this supplementary document.

\noindent\textbf{Training/evaluation flow.}
\begin{enumerate}
\item Resize and normalize each input image; load an auxiliary patch mask if the image has a valid overlapping segmentation mask.
\item Extract frozen DINOv2 ViT-B/14 patch embeddings and last-block attention.
\item Compute LATS scores from patch embeddings, attention entropy, feature norm, and local contrast.
\item Apply spatial smoothing, then select $K=\max(K_{\min},\lfloor\beta N\rfloor)$ patches using straight-through Gumbel top-$K$ during training or deterministic top-$K$ at evaluation.
\item Mean-pool retained patch embeddings and classify with the MLP head.
\item Optimize classification, diversity, lesion-localization, and attention-distillation losses; the budget term records integer/min-token deviation and does not provide scorer supervision.
\item For budget sweeps, evaluate LATS and independently trained baseline heads at $\beta \in \{0.1,\ldots,1.0\}$.
\end{enumerate}

\noindent\textbf{Hyperparameter selection.}
The reported weights and curriculum were fixed before the headline robustness runs. They were chosen from pilot runs for stable low-budget training and non-collapsed score maps rather than by exhaustive search. Validation segmentation masks were not used for hyperparameter selection; they are reserved for retention reporting on the 155-image mask-overlap subset.

\noindent\textbf{Result traceability.}
The main paper reports representative budgets for readability; Tables~\ref{tab:supp_isic_sweep} and~\ref{tab:supp_kvasir_sweep} provide the complete ten-budget sweeps, while Tables~\ref{tab:supp_3seed},~\ref{tab:supp_maskfree}, and~\ref{tab:supp_paired_deltas} provide the aligned robustness and diagnostic control summaries. The code package exposes the same command-line modes used to regenerate full training runs, sweeps, ablations, and mask-free controls from the public datasets.

\subsection{Retention Metric Details}

For ISIC retention evaluation, each available ISIC 2018 segmentation mask is converted to grayscale, resized to the $16\times16$ DINOv2 patch grid, converted to a tensor, and thresholded at $>0.5$ to obtain binary patch labels. The mask is flattened in raster token order to align with the retained-token mask. For a method with one-to-one retained patches, lesion retention is:
\begin{equation}
\mathrm{Retention} =
\frac{\sum_i r_i m_i}{\sum_i m_i},
\end{equation}
where $r_i$ is the retained-patch indicator and $m_i$ is the binary lesion-patch label. Images without valid overlapping segmentation masks are excluded from retention aggregation. In the reported ISIC validation split, 155 images have valid overlapping masks. Validation masks are used only for retention evaluation, not for training, early stopping, or hyperparameter selection.

For methods with one-to-one retained patches, the code also reports lesion precision,
\begin{equation}
\mathrm{Precision} =
\frac{\sum_i r_i m_i}{\sum_i r_i},
\end{equation}
and lesion enrichment, defined as precision divided by lesion-patch prevalence. These are diagnostic complements to retention. The main paper emphasizes retention because it is consistently available for the headline LATS and Random comparison; ToMe merges tokens and therefore does not yield a direct retained-patch mask.

\begin{table*}[t]
\centering
\begin{tabular*}{\textwidth}{@{\extracolsep{\fill}}lrrrrrrrrr}
\toprule
Subset & Images & MEL & NV & BCC & AK & BKL & DF & VASC & SCC \\
\midrule
Full validation & 5067 & 910 & 2568 & 673 & 167 & 516 & 45 & 60 & 128 \\
Mask-overlap validation & 155 & 35 & 120 & 0 & 0 & 0 & 0 & 0 & 0 \\
Mask-overlap training & 617 & 129 & 488 & 0 & 0 & 0 & 0 & 0 & 0 \\
\bottomrule
\end{tabular*}
\caption{ISIC class distribution for the full validation split and the ISIC 2018 mask-overlap subsets used for auxiliary training supervision and retention evaluation. Counts are derived from the experiment validation manifest and official ISIC 2019 labels. The mask-overlap subset is class-limited (MEL/NV), which is why the main paper treats retention as a focused evidence-preservation analysis rather than a full-class validation metric.}
\label{tab:supp_subset}
\end{table*}

\begin{table}[tbp]
\centering
\begin{tabular}{lcc}
\toprule
Statistic & Lesion patches & Prevalence \\
\midrule
Minimum & 6 & 2.3\% \\
25th percentile & 35 & 13.7\% \\
Median & 71 & 27.7\% \\
75th percentile & 104 & 40.6\% \\
Maximum & 248 & 96.9\% \\
Mean $\pm$ std. & $81.1{\pm}58.3$ & $31.6{\pm}22.8\%$ \\
\bottomrule
\end{tabular}
\caption{Lesion-patch prevalence after resizing the 155 validation masks to the $16\times16$ token grid.}
\label{tab:supp_prevalence}
\end{table}

\subsection{Approximate Qualitative Retention Model}

Let there be $N$ patches, with $|\mathcal{P}| = \rho N$ lesion patches. Scores are independent with $s_i \sim \mathcal{N}(s^{+}, \sigma^2)$ for $i \in \mathcal{P}$ and $s_i \sim \mathcal{N}(s^{-}, \sigma^2)$ for $i \notin \mathcal{P}$, where $\Delta = s^{+} - s^{-} > 0$.

The implementation uses $K=\max(K_{\min},\lfloor\beta N\rfloor)$ with $K_{\min}=16$. Ignoring the minimum-token constraint for analysis, top-$K$ selection with $K \approx \beta N$ retains a patch iff its score exceeds the $(1-\beta)$-quantile of the mixture distribution $\theta_\beta$. This Gaussian approximation is intended as a qualitative model of the retention mechanism, not as a formal guarantee. With mixture mean $\mu_{\text{mix}} = s^{-} + \rho\Delta$:
\begin{align}
\theta_\beta &\approx \mu_{\text{mix}} + \sigma \cdot \Phi^{-1}(1-\beta), \\
\mathbb{P}(s_i \leq \theta_\beta \mid i \in \mathcal{P})
&= \Phi\!\left(\Phi^{-1}(1-\beta) - \frac{(1-\rho)\Delta}{\sigma}\right).
\end{align}
Hence:
\begin{equation}
p_{\text{ret}}(\beta) \approx
1 - \Phi\!\left(\Phi^{-1}(1-\beta) - \frac{(1-\rho)\Delta}{\sigma}\right)
\end{equation}

For $\Delta = 0$, $\mathbb{P} \approx 1 - \Phi(\Phi^{-1}(1-\beta)) = \beta$, matching random selection. For $\rho \ll 1$, $(1-\rho) \approx 1$, yielding the simplified form in the main text. This approximation suggests three empirical tendencies: (1) LATS should have its clearest advantage over Random at low budgets (since $\Phi^{-1}(1-\beta)$ grows as $\beta \to 0$); (2) multi-signal scoring can improve retention by reducing effective score noise; and (3) attention distillation can help when it increases separation between lesion-relevant and background scores.

\subsection{Complete Budget Sweeps}

Tables~\ref{tab:supp_isic_sweep} and~\ref{tab:supp_kvasir_sweep} report the complete ten-budget accuracy sweeps used to select the representative budgets shown in the main paper. ISIC is the primary lesion-preservation benchmark because it supports mask-based retention evaluation. Kvasir is an auxiliary accuracy-only dataset: it has no segmentation masks and mixes localized findings with anatomical landmark classes, so it is not used as evidence for the lesion-retention claim.

\begin{table*}[t]
\centering
\begin{tabular*}{\textwidth}{@{\extracolsep{\fill}}lcccccccccc}
\toprule
Method & 0.1 & 0.2 & 0.3 & 0.4 & 0.5 & 0.6 & 0.7 & 0.8 & 0.9 & 1.0 \\
\midrule
NoPruning & 77.5 & 77.5 & 77.5 & 77.5 & 77.5 & 77.5 & 77.5 & 77.5 & 77.5 & 77.5 \\
Random & 76.2 & 77.0 & 76.8 & 77.0 & 77.1 & 76.9 & 76.9 & 77.1 & 77.1 & 77.0 \\
NormBased & 69.7 & 71.8 & 73.1 & 74.0 & 74.5 & 74.9 & 75.4 & 75.3 & 75.5 & 75.7 \\
ToMe & 75.2 & 76.5 & 76.9 & 77.0 & 77.0 & 76.8 & 77.0 & 76.9 & 77.0 & 77.0 \\
AttnEnt & 70.9 & 72.9 & 73.9 & 74.5 & 75.2 & 76.1 & 76.5 & 76.7 & 76.6 & 76.8 \\
LocalCtr & 73.9 & 74.8 & 75.7 & 75.9 & 75.8 & 76.1 & 76.2 & 76.4 & 76.4 & 76.5 \\
\textbf{LATS} & \textbf{77.5} & \textbf{80.1} & \textbf{80.2} & \textbf{79.9} & \textbf{79.3} & 78.6 & 77.9 & 77.1 & 76.1 & 75.1 \\
\bottomrule
\end{tabular*}
\caption{Complete budget sweep: accuracy (\%) on ISIC 2019.}
\label{tab:supp_isic_sweep}
\end{table*}

\begin{table*}[t]
\centering
\begin{tabular*}{\textwidth}{@{\extracolsep{\fill}}lcccccccccc}
\toprule
Method & 0.1 & 0.2 & 0.3 & 0.4 & 0.5 & 0.6 & 0.7 & 0.8 & 0.9 & 1.0 \\
\midrule
NoPruning & 91.6 & 91.6 & 91.6 & 91.6 & 91.6 & 91.6 & 91.6 & 91.6 & 91.6 & 91.6 \\
Random & 90.2 & 91.6 & 90.9 & 90.8 & 91.4 & 91.2 & 91.4 & 91.5 & 91.4 & 91.4 \\
NormBased & 84.1 & 86.2 & 87.9 & 89.8 & 90.4 & 90.4 & 90.3 & 90.4 & 90.4 & 90.4 \\
ToMe & 89.2 & \textbf{92.0} & 91.8 & \textbf{91.9} & 91.6 & 91.6 & 91.5 & 91.2 & 91.2 & 91.4 \\
AttnEnt & 83.2 & 85.6 & 88.1 & 90.0 & 89.8 & 90.5 & 90.9 & 91.3 & 91.2 & \textbf{91.6} \\
LocalCtr & 87.3 & 89.4 & 89.9 & 90.4 & 90.6 & 90.4 & 91.1 & 90.8 & 91.0 & 90.8 \\
\textbf{LATS} & 86.7 & 89.8 & 90.4 & 91.4 & \textbf{91.9} & 91.1 & 91.1 & 90.7 & 90.1 & 88.9 \\
\bottomrule
\end{tabular*}
\caption{Auxiliary accuracy-only sweep on Kvasir v2.}
\label{tab:supp_kvasir_sweep}
\end{table*}

\subsection{Three-Seed Robustness}

Table~\ref{tab:supp_3seed} reports the aligned ISIC robustness check over seeds 42, 43, and 44. Retention is computed for LATS and Random on the 155 validation images that overlap with valid ISIC 2018 segmentation masks. ToMe retention is not reported because merged tokens do not preserve a one-to-one retained-patch set.

\begin{table*}[t]
\centering
\begin{tabular*}{\textwidth}{@{\extracolsep{\fill}}llcccc}
\toprule
$\beta$ & Method & Acc. (\%) & Macro F1 (\%) & Bal. Acc. (\%) & Ret. (\%) \\
\midrule
0.1 & LATS & $77.77{\pm}0.27$ & $67.08{\pm}0.25$ & $65.89{\pm}0.49$ & $28.9{\pm}0.1$ \\
0.1 & Random & $75.75{\pm}0.44$ & $59.50{\pm}1.34$ & $55.39{\pm}1.31$ & $9.9{\pm}0.1$ \\
0.1 & ToMe & $75.94{\pm}0.61$ & $60.96{\pm}1.56$ & $56.90{\pm}1.57$ & --- \\
0.3 & LATS & $80.38{\pm}0.18$ & $69.17{\pm}1.28$ & $66.40{\pm}0.91$ & $67.4{\pm}0.2$ \\
0.3 & Random & $76.57{\pm}0.30$ & $60.95{\pm}1.06$ & $56.72{\pm}1.81$ & $29.7{\pm}0.3$ \\
0.3 & ToMe & $76.93{\pm}0.03$ & $61.92{\pm}0.23$ & $57.81{\pm}0.48$ & --- \\
0.5 & LATS & $79.78{\pm}0.51$ & $67.61{\pm}1.83$ & $64.14{\pm}1.67$ & $85.8{\pm}0.2$ \\
0.5 & Random & $76.74{\pm}0.35$ & $61.68{\pm}0.72$ & $57.15{\pm}1.20$ & $49.9{\pm}0.2$ \\
0.5 & ToMe & $77.13{\pm}0.37$ & $62.82{\pm}0.77$ & $58.73{\pm}0.95$ & --- \\
\bottomrule
\end{tabular*}
\caption{Three-seed ISIC robustness at headline budgets.}
\label{tab:supp_3seed}
\end{table*}

\begin{table*}[p]
\centering
\begin{tabular*}{\textwidth}{@{\extracolsep{\fill}}lccccc}
\toprule
$\beta$ & Acc. (\%) & Macro F1 (\%) & Ret. (\%) & Prec. (\%) & Enrich. \\
\midrule
0.1 & $66.5{\pm}3.0$ & $51.2{\pm}3.1$ & $19.5{\pm}14.2$ & $63.1{\pm}46.1$ & $2.00{\pm}1.46$ \\
0.3 & $73.2{\pm}1.9$ & $55.9{\pm}2.7$ & $46.6{\pm}29.1$ & $49.7{\pm}31.0$ & $1.57{\pm}0.98$ \\
0.5 & $74.8{\pm}0.8$ & $56.7{\pm}2.7$ & $62.7{\pm}29.4$ & $39.6{\pm}18.6$ & $1.25{\pm}0.59$ \\
\bottomrule
\end{tabular*}
\caption{Mask-free LATS control over three seeds on ISIC. The architecture, epochs, and budgets match LATS, but $\mathcal{L}_{\text{lesion}}$ is disabled. Precision and enrichment are diagnostic retained-patch metrics computed only where masks are available.}
\label{tab:supp_maskfree}
\end{table*}

\begin{table*}[p]
\centering
\begin{tabular*}{\textwidth}{@{\extracolsep{\fill}}llccc}
\toprule
$\beta$ & Comparison & $\Delta$Acc. & $\Delta$F1 & $\Delta$Ret. \\
\midrule
0.1 & LATS--Random & $2.02{\pm}0.60$ & $7.58{\pm}1.46$ & $19.03{\pm}0.19$ \\
0.3 & LATS--Random & $3.80{\pm}0.35$ & $8.22{\pm}1.46$ & $37.63{\pm}0.24$ \\
0.5 & LATS--Random & $3.05{\pm}0.84$ & $5.93{\pm}2.29$ & $35.82{\pm}0.23$ \\
0.1 & LATS--ToMe & $1.84{\pm}0.38$ & $6.12{\pm}1.44$ & --- \\
0.3 & LATS--ToMe & $3.45{\pm}0.16$ & $7.25{\pm}1.25$ & --- \\
0.5 & LATS--ToMe & $2.66{\pm}0.37$ & $4.79{\pm}1.39$ & --- \\
\bottomrule
\end{tabular*}
\caption{Paired seed-wise mean differences in percentage points over seeds 42, 43, and 44. Retention differences are reported only for methods with one-to-one retained patch masks.}
\label{tab:supp_paired_deltas}
\end{table*}

Mask-free LATS is intentionally diagnostic: its per-seed retention spans 3.1--28.5\% at $\beta=0.1$, 13.1--66.0\% at $\beta=0.3$, and 29.3--85.0\% at $\beta=0.5$, so the mean retention advantage over Random should be read together with its high instability.

\subsection{Additional Baseline Positioning}

\noindent\textbf{Single-signal protocol.}
The main paper reports independently trained single-signal selector baselines rather than inference-time toggles of a full LATS scorer. These protocols answer different questions: independently trained selectors test whether one cue is sufficient as a standalone routing policy, while inference-time toggles test the robustness of a scorer already trained with all cues. We therefore use the independently trained baselines as the primary evidence for the value of combining weak cues.

\noindent\textbf{Prompt-guided medical token pruning.}
PrATo~\cite{dutta2025prato} uses box-prompt spatial priors to rank and prune tokens for medical image segmentation, including ISIC experiments. MedTokenBudget instead studies prompt-free classification and mask-based retention under a post-backbone token budget.

\noindent\textbf{Attribution-based top-$K$ selectors.}
Transformer relevance propagation~\cite{chefer2021transformer} and GradCAM-style maps are useful localization baselines, but they are prediction-conditioned and require a backward relevance pass. The current sweep therefore compares forward-pass routing policies; attribution-based top-$K$ routing is left to future work.

\noindent\textbf{Mask-oracle upper bound.}
An oracle that directly selects the ground-truth lesion-overlap patches would provide an upper bound on retention, but it would use segmentation masks at inference time and is therefore not a deployable prompt-free selector. We treat it as a useful future diagnostic rather than part of the current forward-pass benchmark.

\end{document}